\documentclass[10pt,twocolumn,letterpaper]{article}

\usepackage[pagenumbers]{cvpr} %

\usepackage{graphicx}
\usepackage{amsmath}
\usepackage{amssymb}
\usepackage{booktabs}
\usepackage[usenames,dvipsnames]{xcolor}
\usepackage{ifthen}
\usepackage[percent]{overpic}  %

\usepackage[pagebackref,breaklinks,colorlinks]{hyperref}
\usepackage{multirow}

\usepackage[capitalize]{cleveref}
\crefname{section}{Sec.}{Secs.}
\Crefname{section}{Section}{Sections}
\Crefname{table}{Table}{Tables}
\crefname{table}{Tab.}{Tabs.}
\def\equationautorefname~#1\null{Eq.~(#1)\null} %

\usepackage{array}
\newcolumntype{H}{>{\setbox0=\hbox\bgroup}c<{\egroup}@{}}
\providetoggle{showcomments}
\settoggle{showcomments}{true} 								%

\iftoggle{showcomments}{%

    \newcommand{\todo}[1]{\textcolor{blue}{\textbf{TODO:} #1}}
    \newcommand{\resolved}[3][]{\ifstrequal{#1}{resolved}{\textcolor{blue}{RESOLVED:}~\textbf{{\MakeUppercase #2:}}~{#3}}{\textbf{\MakeUppercase #2:}~#3}}
    \newcommand{\michal}[2][]{\textcolor{ForestGreen}{\resolved[#1]{michal}{#2}}}
    \newcommand{\alina}[2][]{\textcolor{NavyBlue}{\resolved[#1]{alina}{#2}}}
    \newcommand{\vitto}[2][]{\textcolor{red}{\resolved[#1]{vitto}{#2}}}
    \newcommand{\jasper}[2][]{\textcolor{violet}{\resolved[#1]{jasper}{#2}}}
    
    \newcommand{\tm}[2][]{\textcolor{magenta}{\resolved[#1]{TM}{#2}}}
    
}{%

    \newcommand{\todo}[1]{}
    \newcommand{\michal}[2][]{}
    \newcommand{\alina}[2][]{}
    \newcommand{\vitto}[2][]{}
    \newcommand{\jasper}[2][]{}
    \newcommand{\tm}[2][]{}
    
    \newcommand{\xhdr}[1]{{\noindent\bfseries #1}.}
}

\newcommand*\diff{\mathop{}\!\mathrm{d}}

\newcommand{\xhdr}[2][1.8mm]{\vspace{#1}{\noindent\bfseries #2.}\quad}

\begin{document}

\title{Transferability Estimation using Bhattacharyya Class Separability}

\author{
Michal P\'{a}ndy\thanks{Currently at Waymo.} \quad 
Andrea Agostinelli \quad 
Jasper Uijlings \quad
Vittorio Ferrari \quad
Thomas Mensink\\[5mm]
Google Research\\
{\tt\small \{michalpandy, agostinelli, jrru, vittoferrari, mensink\}@google.com}
}

\maketitle

\begin{abstract}
Transfer learning has become a popular method for leveraging pre-trained models in computer vision. However, without performing computationally expensive fine-tuning, it is difficult to quantify which pre-trained source models are suitable for a specific target task, or, conversely, to which tasks a pre-trained source model can be easily adapted to. In this work, we propose Gaussian Bhattacharyya Coefficient (GBC), a novel method for quantifying transferability between a source model and a target dataset.
In a first step we embed all target images in the feature space defined by the source model, and represent them with per-class Gaussians. Then, we estimate their pairwise class separability using the Bhattacharyya coefficient, yielding a simple and effective measure of how well the source model transfers to the target task. We evaluate GBC on image classification tasks in the context of dataset and architecture selection. Further, we also perform experiments on the more complex semantic segmentation transferability estimation task. We demonstrate that GBC outperforms state-of-the-art transferability metrics on most evaluation criteria in the semantic segmentation settings, matches the performance of top methods for dataset transferability in image classification, and performs best on architecture selection problems for image classification.

\end{abstract}

\section{Introduction}

The goal of transfer learning is to reuse knowledge learned on a source task to help train a model for a target task.
Currently, the  most  common  form  of  transfer  learning  in  computer vision is to pre-train a source model on the ILSVRC'12 dataset~\cite{russakovsky15ijcv} and then fine-tune it on the target dataset~\cite{azizpour15pami, chu16eccv, girshick15iccv, he17iccv, huh16nips, kornblith19cvpr, shelhamer16pami, zhou19arxiv}.
However, each target task may benefit from a different source model architecture~\cite{chen18pami, he16cvpr, newell16eccv, ronneberger15miccai} or different source dataset~\cite{mensink21arxiv, ngiam18arxiv, yan20cvpr}.
The challenge then becomes to determine which (pre-trained) source model is most suitable for a particular target task, or to which target task a specific model can be easily adapted. Determining this by fine-tuning all combinations of source models and target datasets is computationally prohibitive.

To address this problem, several recent works introduced transferability metrics~\cite{bao2019hscore, li2021ranking, nguyen2020leep, tan2021otce, tran2019transferability, you2021logme}, which aim at predicting how well a source model transfers to a given target dataset. A good transferability metric is computationally efficient, and its predictions correlate well with the final performance of a model after fine-tuning on the target dataset.
Typically a transferability metric is estimated by applying the source model to the target dataset to extract embeddings or predictions, which are then combined with the target ground-truth labels to measure transferability.

\begin{figure}[t]
    \centering
    \includegraphics[width=\columnwidth]{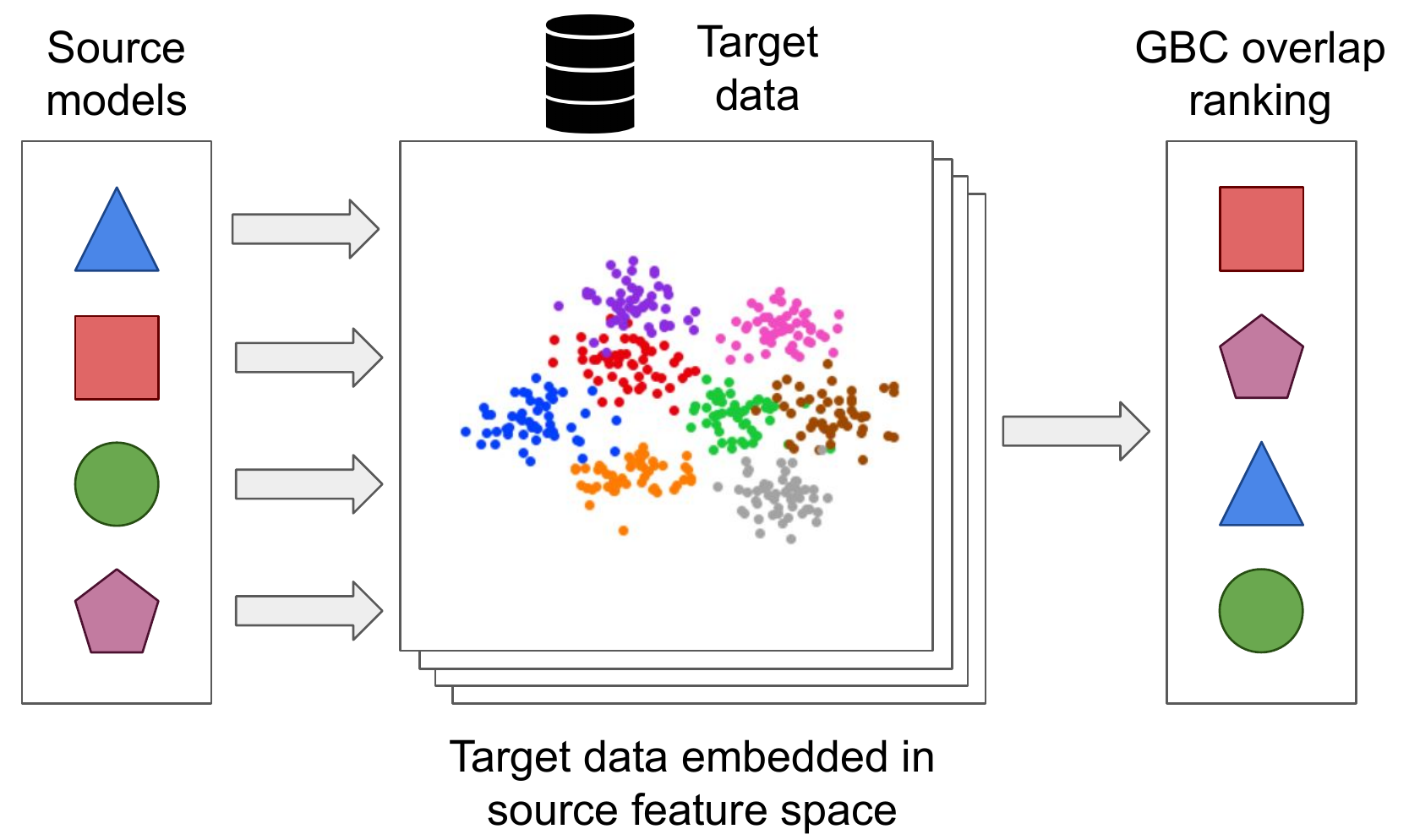}
    \caption{This figure illustrates the high-level overview of our approach. On the left, we use the pre-trained source models to embed data into the source models' feature space. On the right, we use GBC to rank these methods based on how much classes overlap in corresponding embedding spaces.}
    \label{fig:intro-fig}
\end{figure}

This paper proposes a novel transferability metric: the Gaussian Bhattacharyya Coefficient (GBC).
The main idea is to measure the amount of overlap between target classes in the feature space of the source model (Fig.~\ref{fig:intro-fig}). If this overlap is small, the target classes are easily separated which means the knowledge in the source model is useful for the target task and the source model should transfer well. Conversely, if the overlap is large, the target classes are difficult to separate and the source model transfers badly to this target task.
In order to estimate the amount of overlap, we apply the feature extractor of the source model to the target dataset and model each target class as a Gaussian distribution in this space.
Importantly, we carefully apply regularization techniques to ensure that the Gaussian model can accurately represent each class.
Then, we measure the sum of the overlaps between each pair of target classes using the Bhattacharyya coefficient.
The Bhattacharyya coefficient has a closed-form solution when applied on Gaussian distributions.
We use this overlap as our transferability metric.

We perform extensive experiments on two tasks.
First we consider image classification, the primary focus of previous works on transferability metrics~\cite{bao2019hscore, li2021ranking, nguyen2020leep, tan2021otce, tran2019transferability, you2021logme}. Additionally, we consider a realistic transfer learning scenario for the task of semantic segmentation by considering transfer across a large variety of datasets from different image domains.
Our experiments demonstrate that GBC outperforms several state-of-the-art transferability metrics: LEEP~\cite{nguyen2020leep}, LogME~\cite{you2021logme}, H-score~\cite{bao2019hscore}.
Furthermore, we demonstrate that our method is computationally efficient. 

In summary, our paper makes the following contributions:
1) We introduce GBC, a new transferability metric which measures the amount of overlap between target classes in the source feature space. Since we model the target samples with per-class Gaussians, the GBC can be estimated in closed form;
2) We lift transferability experiments to a realistic transfer learning scenario for semantic segmentation;
3) We experimentally demonstrate that our GBC method outperforms other transferability metrics, including LEEP\cite{nguyen2020leep}, LogME\cite{you2021logme}, and H-score~\cite{bao2019hscore}.

\section{Related work}
While our work falls into the broad domain of transfer learning~\cite{pan2009survey, weiss2016survey}, and relates to model selection~\cite{ding2018model, raschka2018model}, and domain adaption~\cite{ben2007analysis, daume2009frustratingly, pan2010domain}, in this section we discuss the most relevant related work in estimating transferability metrics.
We structure these along four general paradigms.

\xhdr{Task relatedness}
Pioneering works in task relatedness~\cite{kifer2004detecting, bendavid2007, mansour2009} introduce symmetric measures between source and target tasks or domains. The intuition is that related tasks could be learned together more efficiently~\cite{bendavid2007}. 
While intuitively task relatedness should correlate with transferability, these particular measures are generally hard to estimate.
Moreover, the relatedness measures are symmetric, while transfer is asymmetrical: ImageNet is probably a very good source dataset for CIFAR-10, while the reverse is likely not true~\cite{mensink21arxiv, nguyen2020leep}.

\xhdr{Label comparison-based methods} 
LEEP~\cite{nguyen2020leep} and NCE~\cite{tran2019transferability} use the labels of the source domain and the target domain to construct transferability metrics.
In NCE by Tran \etal~\cite{tran2019transferability}, they assume that the images of a source and target task are identical, but their labels differ. Then, they use the negative conditional entropy between the target labels and the source labels as a tranferability metric.
Nguyen \etal propose LEEP~\cite{nguyen2020leep}, where the source model is applied to the target dataset. The resulting label predictions are utilised for computing a log-likelihood between the target labels and the source model predictions.
An assumption in label comparison-based models is their dependence on the source output label space. Specifically, two source models with an identical feature extractor yet different classification heads will produce different transferability scores. In contrast source embedding-based methods rely purely on the underlying feature extractor.

\xhdr[1mm]{Source embedding-based methods} 
Source embedding-based methods utilize the embeddings of target samples obtained via a pre-trained source model. 
Target embeddings are used together with their labels to compute various distance metrics. 
Cui \etal~\cite{Cui_2018_CVPR} propose to compute the Earth Mover's Distances between the class conditioned means of the embeddings. 
In H-score by Bao \etal~\cite{bao2019hscore}, high transferability estimates are assigned to sources, where the embeddings display low feature redundancy and high inter-class variance. 
Li \etal~\cite{li2021ranking} introduce $\mathcal{N}$LEEP, an extension to LEEP where the authors fit a Gaussian Mixture Model of the target data in the embedding space and use this in place of the source model's classification head to compute the LEEP score.
Finally, You \etal~\cite{you2021logme} propose the state-of-the-art LogME score which treats each target label as a linear model with Gaussian noise, and then optimise the parameters of the prior distribution to find the average maximum (log) evidence of labels given the target sample embeddings.
Our work also falls into the source embedding-based methods, but we directly consider the separability of class conditioned target embeddings. 

\xhdr{Optimal transport} 
There have been several works proposing transferability estimation based on optimal transport (OT), including~\cite{tan2021otce, melis2020gdd}. 
The underlying assumption is that when in the source model's embedding space the source and target datasets have similar geometrical structures, and hence have a low OT-distance, the given source model is a suitable for the given target dataset.
With~\cite{melis2020gdd}, we share the idea to model classes as Gaussian distributions in the embeddeding space.
However, OT based approaches have some serious drawbacks:
(1) The method in~\cite{tan2021otce} relies on parameter tuning based on ground-truth transferability scores;
(2) These methods require access to the source training set; and
(3) Computing the (regularized) OT distance scales quadratically in the number of data samples, which makes it practically infeasible to compute transferability scores for large datasets (including ImageNet).
\section{Method}
\label{sec:method}
\newcommand{\BC}{\mathrm{BC}}
\newcommand{\GBC}{\mathrm{GBC}}

\subsection{Formal description}
\label{sec:formal-description}
Before we describe our method, we first provide a more formal description of the problem at hand. 
The goal is to estimate the transferability score $\mathcal{S}_{s \rightarrow t}$ of a \textit{source model} $m_s$ for a particular \textit{target task} $t$. The target task $t$ is described by a training set $\mathcal{D}_t$ containing images and ground truth label pairs $(x_t, y_t)$. 

A good transferability metric $\mathcal{S}_{s \rightarrow t}$ correlates with the accuracy $\mathcal{A}_{s \rightarrow t}$ of the target model $m_{s\rightarrow t}$. The accuracy $\mathcal{A}_{s \rightarrow t}$ is measured by evaluating $m_{s\rightarrow t}$ on the (unseen) test set of the target task $\mathcal{D}_t^{\mathrm{test}}$.
To create the target model $m_{s\rightarrow t}$, it is initialized using the weights of the source model $m_s$, after which it is fully fine-tuned on the target task $t$ using the target training set $\mathcal{D}_t$.
However, since fully fine-tuning $m_{s\rightarrow t}$ is computationally expensive, instead we want to predict how it will transfer using a computational efficient transferability metric $\mathcal{S}_{s \rightarrow t}$.

The source model $m_s$ is defined by (a) the network architecture, such as ResNet50~\cite{he16cvpr} or VGG16~\cite{simonyan15iclr}; and (b) the training dataset used to train the source network, such as supervised classification on ImageNet~\cite{russakovsky15ijcv}.

For our method, we assume that we have access to the image embedding function of the source model $f_s(x)$, which returns a feature vector representation of image $x$.
Our method only relies on the feature extractor $f_s(x)$, similar to H-score\cite{bao2019hscore} and LogME\cite{you2021logme}.
In contrast, optimal transport based methods require access to the source (training) dataset $\mathcal{D}_s$~\cite{melis2020gdd, tan2021otce}, and LEEP requires the per target example predictions in the source label space~\cite{nguyen2020leep}. 

\xhdr{Evaluating transferability}
We evaluate the performance of the transferability metrics by evaluating the correlation between $\mathcal{S}_{s \rightarrow t}$ and $\mathcal{A}_{s \rightarrow t}$, as measured by the weighted Kendall tau rank correlation $\tau_w$, as proposed by~\cite{you2021logme}.

In contrast to the Pearson $r$ correlation coefficient which measures a linear relation between $\mathcal{S}$ and $\mathcal{A}$, the Kendall rank correlation allows for highly non-linear relations, since it correlates rankings.
The weighted Kendall correlation $\tau_w$ places higher weights on the models with the highest accuracies. 
This incorporates the rationale that it is more important to have the top few models correctly ranked, than the models with lower accuracies. For a more elaborate discussion on the appropriateness of the weighted Kendall tau, we refer the reader to You \etal \cite{you2021logme}.

We evaluate $\tau_w$ for different kinds of transferability scenarios, either fixing the target task to find the most suitable source model, or by correlating a fixed source model with different target tasks.

\begin{figure}[t]
    \centering
    \fbox{\includegraphics[width=0.465\columnwidth]{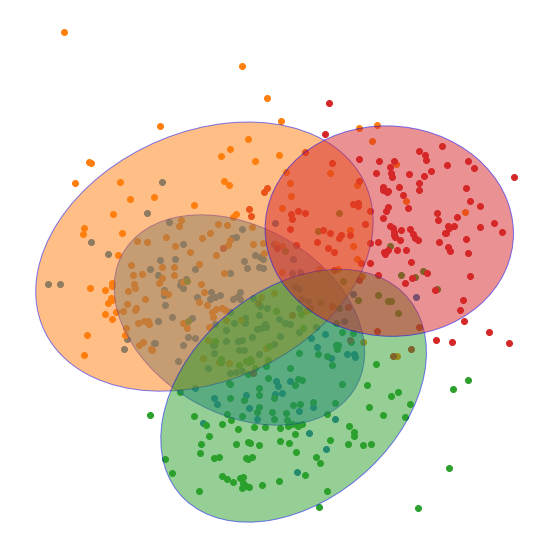}}~%
    \fbox{\includegraphics[width=0.465\columnwidth]{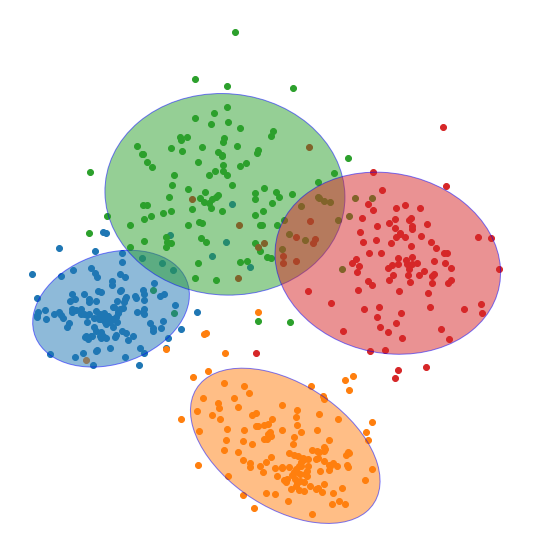}}
    \caption{
    Illustration of the intuition: 
    when images of the target classes overlap less in the embedding space of a source model, then it is a more suitable source to transfer from (left: a poor source; right: a better one).
    We use the Bhattacharyya coefficient to estimate the overlap between classes, each modeled as a Gaussian.
    }
    \label{fig:overlap_intuition}
    \vspace{-5mm}
\end{figure}

\subsection{Class separability}
The key idea behind our method is that if the target images are class-wise separable in the source model feature space $f_s(\cdot)$, then this source model allows for good classification for the target task.
This intuition is shown in \autoref{fig:overlap_intuition}, where we show two embeddings of 4 target classes. We argue that the left dataset is more difficult to transfer to than the right dataset because the amount of class overlap is higher.
We posit that the class overlap is proportional to the error of a sufficiently expressive fine-tuned classifier on the target dataset, and hence is proportional with the transferability of the source model to the target data. 
Our approach measures the amount of class separability of the target dataset in the source model feature space and uses that as transferability score $\mathcal{S}_{s \rightarrow t}$.

\begin{figure*}[t]
\hspace{-0.6cm}
\includegraphics[width=1.07\textwidth]{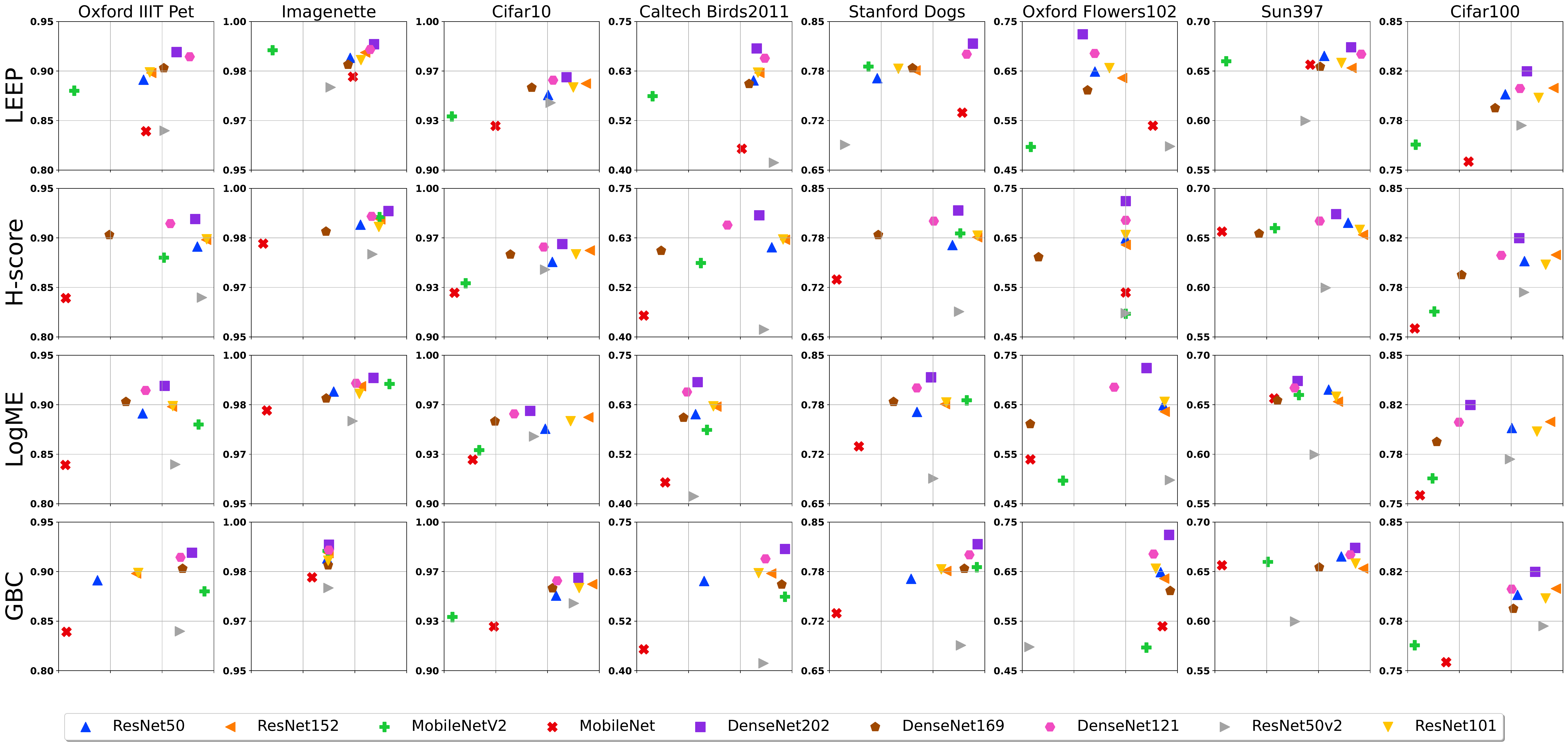}
\caption{
Overview of source selection experiments. For eight different target datasets we show the correlation between the accuracy of the model ($\mathcal{A}$, Y-axis) and the transferability scores ($\mathcal{S}, $X-axis) of LEEP, H-score, LogMe and GBC. See text for details.
}
\label{fig:source_scatter}
\end{figure*}

\xhdr{Bhattacharyya coefficient}
The Bhattacharyya coefficient \cite{bhattacharyya1946measure} ($\BC$) is a measure of the amount of overlap between two distributions; in our case we want to measure the overlap between the probability densities of two target classes $p_{c_i}$, $p_{c_j}$: 
\begin{equation}
    \label{eq:bhat}
    \BC(p_{c_i}, p_{c_j}) = \int \sqrt{p_{c_i}(x) \, p_{c_j}(x)} \diff x
\end{equation}

\xhdr{Per-class Gaussian distributions}
In order to compute the Bhattacharyya coefficient, we need to define the probabilistic model $p_c$ for the target classes.
We chose to model each class distribution with a Gaussian in the source embedding space: $p_c = \mathcal{N}(\mu_c, \Sigma_c)$, with:
\begin{align*}
    \mu_c &= \frac{1}{N_c} \sum_i [\![y_i\!=\!c]\!] \, f_s(x_i)\\
    \Sigma_c &= \frac{1}{N_c - 1} \sum_i [\![y_i\!=\!c]\!] \, (f_s(x_i) - \mu_c) \, (f_s(x_i) - \mu_c)^\top
\end{align*}
where $N_c=\sum_i [\![y_i\!=\!c]\!]$, the number of images in this class. 

While we do not suggest that per-class distributions are necessarily (multivariate) Gaussians, 
the advantage of using this model is 
that the Bhattacharyya coefficient can be computed in closed form from the class means and covariance matrices, using the Bhattacharyya distance $D_B$:
\begin{align}
    D_B(c_i, c_j) &= \frac{1}{8} (\mu_{c_i} - \mu_{c_j})^{\top}\Sigma^{-1}(\mu_{c_i} - \mu_{c_j})\nonumber\\
                &\quad\quad\quad + \frac{1}{2}\text{ln}\left(\frac{|\Sigma|}{\sqrt{|\Sigma_{c_i}||\Sigma_{c_j}|}}\right)\label{eq:bhat_gauss}
\end{align}
where $\Sigma = \tfrac{1}{2}(\Sigma_{c_i} + \Sigma_{c_j})$, the Bhattacharyya coefficient is then $\BC(\cdot, \cdot)=\exp -D_B(\cdot, \cdot)$.

\xhdr{Gaussian Bhattacharyya coefficient}
Our final transferability score estimates the overlap of all classes by taking the sum of the pairwise coefficients:
\begin{equation}
 \GBC_{s\rightarrow t} = - \sum_{i, j} \, [\![i \neq j]\!] \, \BC(c_i, c_j)
\label{eq:bhat_pw_sum}
\end{equation}
The final score uses the \textit{negative} sum, because higher Bhattacharyya coefficients correspond to more overlap between the classes and therefore less transferability.

\xhdr{Theoretical guarantee} %
A nice property of GBC is that it provides some theoretical guarantee: 
when a 
classification head is fine-tuned on top of a fixed feature backbone
and the per-class Gaussian assumption holds,
then GBC is equivalent to an upper-bound on the optimal Bayes classification error \cite{mak1996phone, fukunaga2013introduction}. %
However, when the full model is fine-tuned, it is difficult to draw such a strong guarantee, as for all transferability metrics. For this, we rely on strong empirical results to demonstrate that GBC works well also in this general case.

\begin{table*}[!htpb]

\centering
\resizebox{.99\textwidth}{!}{%

\begin{tabular}{ccccccccc|c} \toprule
& Pets & Imagenette& CIFAR-10& CUB'11& Dogs & Flowers102& SUN & CIFAR-100 &\textbf{Average}\\ \midrule
LogMe &-0.06& 0.58& 0.25& 0.2& 0.08& 0.00& -0.19& 0.34 &0.15\\
H-score &0.06& 0.59& 0.45& 0.16& -0.01& 0.09& 0.09& 0.34 &0.22\\
LEEP &\textbf{0.63}& \textbf{0.65}& \textbf{0.52}& 0.25& 0.59& -0.46& \textbf{0.40} & \textbf{0.55} &0.39 \\
GBC &0.55& 0.63& 0.46& \textbf{0.43}& \textbf{0.80}& \textbf{0.23}& 0.32& 0.35 &\textbf{0.47}\\ \bottomrule
\end{tabular}
}
\caption{
Overview of results for transferability for source selection in  image  classification.
We depict for eight different target datasets the weighted Kendall $\tau_w$ between the accuracy of the fine-tuned model and the transferability scores from LEEP, LogMe, H-score and the proposed GBC. Our proposed method obtains the highest average $\tau_w$ across the different datasets. 
}\vspace{-3mm}
\label{tab:source_selection_full}
\end{table*}%

\subsection{Practical considerations}
\label{sec:practicalconsideration}
\xhdr{PCA dimensionality reduction}
In practice, we transform the source embedding using the PCA projection into a fixed dimensional feature space of 64 dimensions. 
The reason for doing so is that different source architectures produce features with different number of dimensions and the Bhattacharyya coefficient is affected by the dimensionality due to its use of the determinant of the covariance matrix (Eq.~\eqref{eq:bhat_gauss}), which would make GBC scores difficult to compare.
Moreover, reducing the number of dimensions allows to better estimate the Gaussian model.

\xhdr{Covariance estimation}
To compute GBC, we need to estimate the per-class covariance matrices for all target classes.
However, estimating the full covariance is infeasible, since the number of samples in a class can be very low, for example in the Caltech-USCD Birds\cite{CUB2011} dataset, on average there are only $30$ samples per class.
Therefore, we experiment with both diagonal covariance matrices and spherical ones.

\xhdr{Time complexity}
In order to compute GBC, we first extract source model features $f_x(\cdot)$ for all images in the target data, the corresponding complexity if $O(N F)$, for $N$ images, and where $F$ denotes the complexity of extracting features.
First, for PCA estimation using SVD, it costs $O(N D^2)$ to obtain the projection matrix.
Then, to project samples and obtain their per-class means and covariance estimates in the reduced $d$-dimensional space is $O(NDd)$ and $O(NDd^2)$, respectively. 
Finally, computing the Bhattacharyya distance \eqref{eq:bhat_gauss} between two classes in the reduced space with diagonal covariance matrices costs $O(d)$, so estimating our transferability metric \eqref{eq:bhat_pw_sum} is $O(C^2d)$.
In practice, the total run time largely depends on the cost of extracting features for the target dataset: $O(NF)$.
\section{Experiments}
\label{sec:experiments}

In this section, we evaluate our proposed GBC transferability metric. 
We consider various transfer learning tasks to compare our proposed method against related work. 

\subsection{Classification: architecture transferability}
\label{sec:source-selection}

\xhdr{Experimental setup}
We consider several different source model architectures pre-trained on ImageNet. We want to identify which architecture would perform best on a given target dataset. To this end, we follow the experimental setup in~\cite{you2021logme} and evaluate our method using $8$ different target datasets and $9$ commonly utilized network architectures.
Concretely, we fix the target dataset and we compute $\mathcal{A}_{s \rightarrow t}$ and $\mathcal{S}_{s \rightarrow t}$ for each architecture. Then, we measure weighted Kendall rank correlation $\tau_w$ (as proposed by You  \etal \cite{you2021logme}) between the reference $\mathcal{A}_{s \rightarrow t}$ and predicted $\mathcal{S}_{s \rightarrow t}$ across all the architectures, and report the results.
We repeat this experiment for every target dataset.

We use the following target datasets: CIFAR-10 \& 100~\cite{krizhevsky09}, Imagenette~\cite{imagenette}, Oxford IIIT Pets~\cite{parkhi12cvpr}, Caltech-USCD Birds 2011 (CUB'11)~\cite{CUB2011}, Stanford Dogs~\cite{khosla2011dogs}, Oxford Flowers 102~\cite{nilsback08ieee}, and SUN-397~\cite{xiao10cvpr}. 

As source architectures we use ResNet-50, ResNet-101 \& ResNet-152~\cite{he16cvpr}, ResNetV2-50~\cite{he2016identity}, DenseNet-101, DenseNet-169 \& DenseNet-201~\cite{huang2017densely}, MobileNet~\cite{howard2017mobilenets}, and MobileNetV2~\cite{sandler18cvpr}
from the Keras library~\cite{chollet2015keras_applications}.

The target accuracy $\mathcal{A}_{s \rightarrow t}$ is computed by evaluating the target model after fine-tuning each architecture on each target dataset for 100 epochs (with SGD with Momentum, using a batch size of $64$ and learning rate of $10^{-4}$).

We compare our method to three competitive baselines: H-score~\cite{bao2019hscore}, LEEP~\cite{nguyen2020leep} and LogME~\cite{you2021logme}\footnote{LEEP \& LogMe: \href{https://github.com/thuml/LogME}{github.com/thuml/LogME}; H-score: \href{https://git.io/J1WOr}{git.io/J1WOr}}.

\newlength{\mywidth}
\setlength{\mywidth}{.11\textwidth}
\newcommand{\gbcfigure}[3][]{%
    \fbox{%
    \begin{overpic}[width=\mywidth]{#2}
        \put (0, 90) {{\footnotesize $\textrm{gbc}\!:\!#3$}}
        \ifthenelse { \equal {#1} {} }{}{
            \put (-15, 50){\makebox(0,0){\rotatebox{90}{\footnotesize #1}}}
        }
    \end{overpic}}
}
\newcommand{\accfigure}[2]{%
    \fbox{%
    \begin{overpic}[width=\mywidth]{#1}
        \put (50, 90) {{\footnotesize $\textrm{acc}\!:\!#2$}}
    \end{overpic}}
}

\begin{figure}[t]
    \centering
    \begin{tabular}{@{}c@{}c@{}c@{}c@{}}
    \multicolumn{2}{c}{\textbf{DenseNet}}& \multicolumn{2}{c}{\textbf{MobileNet}}\\
    \gbcfigure[CIFAR-10]{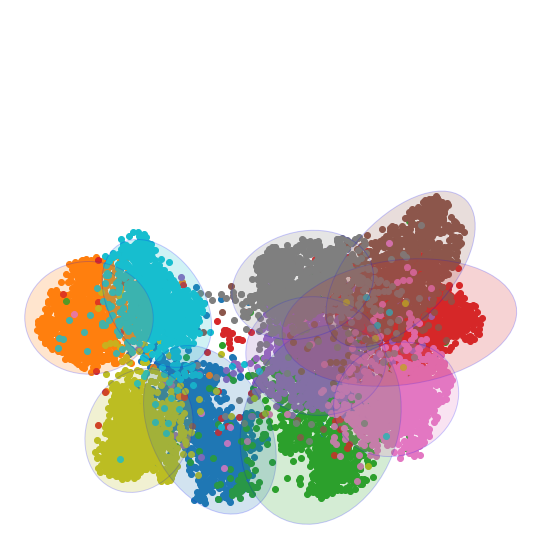}{-1.1}&
    \accfigure{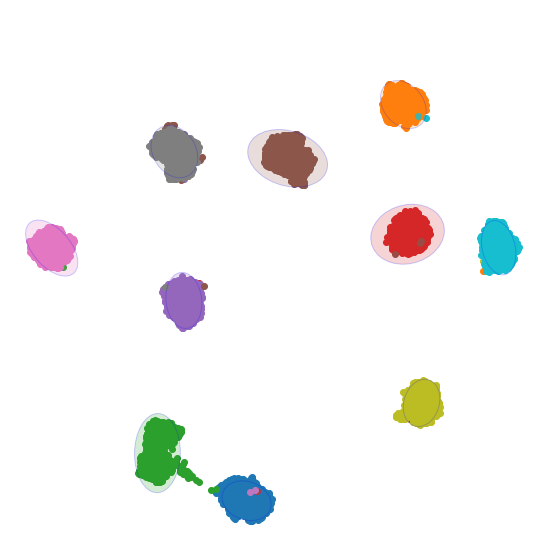}{.96}&
    \gbcfigure{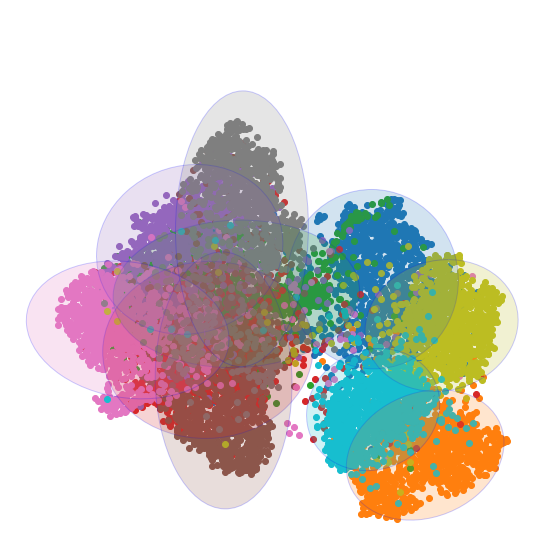}{-2.8}&
    \accfigure{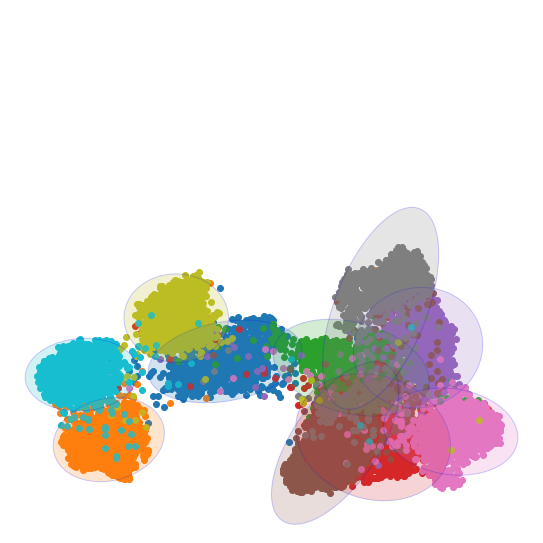}{.93}\\[0mm]
    \gbcfigure[CUB'11]{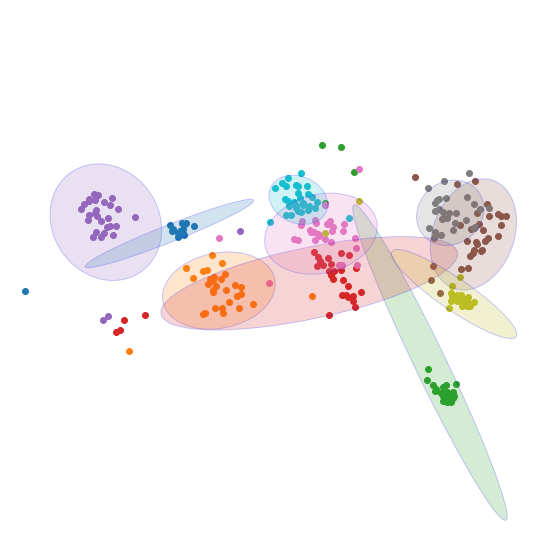}{-159}&
    \accfigure{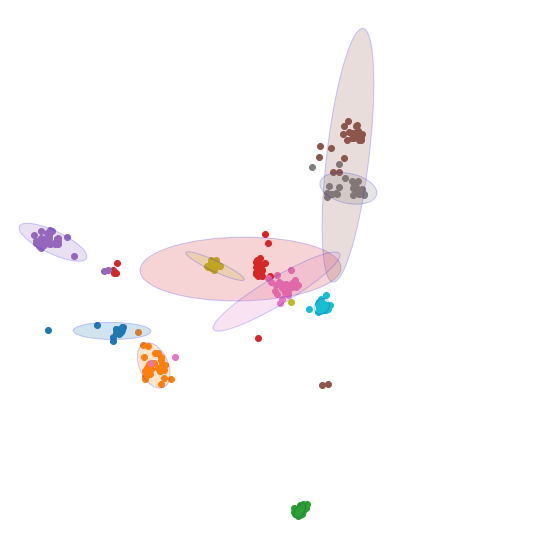}{.68}&
    \gbcfigure{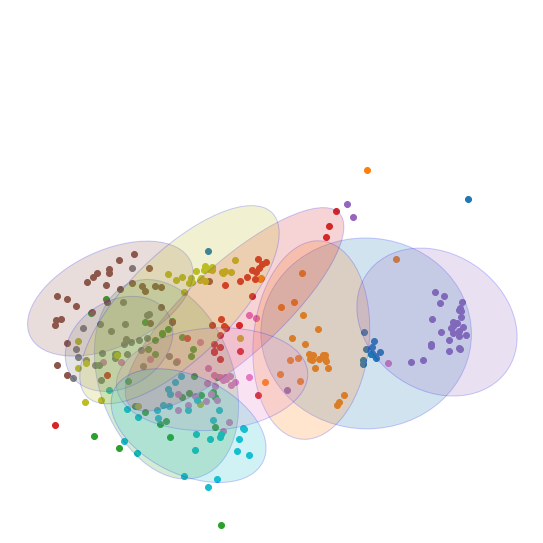}{-419}&
    \accfigure{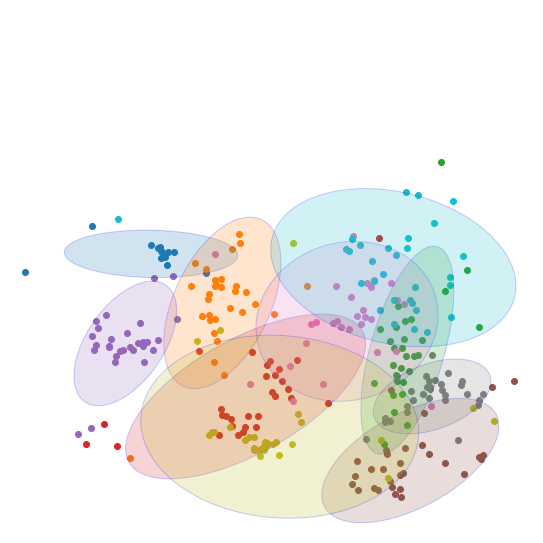}{.45}\\[0mm]
    Before & After & Before & After\\
    \end{tabular}%
    \caption{Feature distribution of CIFAR-10 (top) %
    and 10 (randomly selected) classes of CUB'11 (bottom), visualized with UMAP. 
    }\label{fig:overlap}\vspace{-5mm}
\end{figure}

\xhdr{Main results and comparisons}
We present the results of the full source architecture transferability experiment in \autoref{tab:source_selection_full} (see more results in ~\autoref{app:image_classification}). 
Notably, GBC achieves the highest average rank correlation $\tau_w$ of $.47 $ over all the target datasets. Moreover, GBC is the only method to exhibit positive rank correlations for all target datasets.
GBC determines the single best performing architecture for $3$ target datasets, while LEEP and H-score for $2$, and LogME for none.
Furthermore, the best architecture is among the top-$3$ suggested models by GBC in $7$ datasets, while only in $6$ for LEEP, in $3$ for H-score, and in $1$ for LogME.

\autoref{fig:source_scatter} presents the scatter plots of the accuracies $\mathcal{A}$ and estimated transferability scores $\mathcal{S}$ obtained by each method on each dataset. GBC showcases increasing trends across all datasets.
These results demonstrate that GBC outperforms previous work for source architecture selection.

\autoref{fig:overlap} shows the feature distributions before and after fine-tuning for two models with different GBC transferability scores (DenseNet and MobilNet).
Each row shows a separate experiment on a different target dataset (CIFAR-10, CUB'11).
In both cases, MobileNet has lower GBC scores than DenseNet and also results in lower accuracy after fine-tuning, demonstrating that our method works as intended.

\xhdr{Influence of regularization}
We evaluated the influence of GBC's regularization parameters (\autoref{sec:practicalconsideration}).
We used CIFAR-10 as the target dataset and transferred from the 9 source architectures listed above. For PCA we considered \{16,32,64,128\}-dimensional projections, and for the covariance estimation the regularization variants: \{full, diagonal, spherical\}. 
From the results we conclude that spherical regularization with 64-dimensional PCA projections delivers the best performance. 
Please see \autoref{app:ablations} for full details.
Hence, in all classification experiments we use these settings (\autoref{sec:source-selection} \& \autoref{sec:target-selection}).

We want to highlight that using these settings the covariance estimation is robust, even in a low data regime:
(1) On the smallest dataset we consider (CUB'11, 29 samples per class), GBC outperforms all previous methods in \autoref{tab:source_selection_full} and is on-par with the best in \autoref{tab:target-selection};
(2) We computed the Pearson correlation ($\rho$) between GBC's performance and the number of samples per class. 
They are essentially uncorrelated ($\rho =-0.048$), suggesting that GBC does not perform worse with fewer samples per class.

\xhdr{Computational cost}
To provide an indicative reference, we compare here the run times of several transferability metrics on CIFAR-100 (on a single CPU).
After the feature extraction stage (shared by all metrics), GBC runs in 7.8s, vs 12.0s for LogME, 6.1s for H-score, and 0.2s for LEEP.

\newcommand{\cifar}{CIFAR-10}%
\newcommand{\cifarh}{CIFAR-100}%
\newcommand{\fashion}{F-MNIST}
\newcommand{\birds}{CUB'11}
\newcommand{\sun}{SUN}

\begin{table}[t]
\centering
\resizebox{0.98\columnwidth}{!}{%

\begin{tabular}{cHcccc} \toprule
 \textbf{Source} & \textbf{Target dataset}  & \textbf{LEEP}         & \textbf{LogME}        & \textbf{H-score}      & \textbf{GBC}\\ 
                 &                          & \cite{nguyen2020leep} & \cite{you2021logme}   & \cite{bao2019hscore}  & \textit{Ours}\\\midrule
&&\multicolumn{4}{c}{\textbf{CIFAR-10}}\\
\birds &\cifar &0.68 &0.71 &0.69 &\textbf{0.72} \\
\cifarh &\cifar &\textbf{0.75} &\textbf{0.75} &0.73 &0.69 \\
\fashion &\cifar &0.68 &0.70 &\textbf{0.72} &0.68 \\
\sun &\cifar &0.73 &\textbf{0.75} &0.72 &0.67 \\ 
ImageNet &\cifar &0.68 &0.68 &0.69 &\textbf{0.71} \\\midrule
&&\multicolumn{4}{c}{\textbf{CIFAR-100}}\\
\birds &\cifarh &\textbf{0.90} &0.29 &0.59 &\textbf{0.90} \\
\cifar &\cifarh &\textbf{0.92} &0.29 &0.88 &\textbf{0.92} \\
\fashion &\cifarh &\textbf{0.88} &0.24 &0.26 &\textbf{0.88} \\
\sun &\cifarh &\textbf{0.90} &0.30 &0.88 &\textbf{0.90} \\ 
ImageNet &\cifarh &0.91 &0.25 &0.88 &\textbf{0.92} \\\midrule
&&\multicolumn{4}{c}{\textbf{Fashion-MNIST}}\\
\birds &\fashion &\textbf{0.72} &0.71 &0.71 &0.71 \\
\cifar &\fashion &0.72 &\textbf{0.73} &0.69 &0.69 \\
\cifarh &\fashion &\textbf{0.71} &\textbf{0.71} &0.70 &0.69 \\
\sun &\fashion &\textbf{0.71} &\textbf{0.71} &0.69 &\textbf{0.71} \\ 
ImageNet &\fashion &\textbf{0.72} &0.71 &0.69 &0.70 \\\midrule
&&\multicolumn{4}{c}{\textbf{Caltech-USCD Birds 2011}}\\
\cifar &\birds &\textbf{0.87} &-0.59 &0.83 &0.86 \\
\cifarh &\birds &\textbf{0.87} &-0.58 &0.80 &\textbf{0.87} \\
\fashion &\birds &\textbf{0.70} &-0.50 &0.51 &0.69 \\
\sun &\birds &\textbf{0.88} &-0.60 &0.80 &\textbf{0.88} \\
ImageNet &\birds &\textbf{0.89} &-0.59 &0.73 &0.88 \\\midrule
&&\multicolumn{4}{c}{\textbf{SUN-397}}\\
\birds &\sun &\textbf{0.95} &0.87 &0.54 &\textbf{0.95} \\
\cifar &\sun &\textbf{0.95} &0.87 &0.12 &\textbf{0.95} \\
\cifarh &\sun &\textbf{0.95} &0.88 &0.51 &\textbf{0.95} \\
\fashion &\sun &\textbf{0.95} &0.86 &0.54 &\textbf{0.95} \\
ImageNet &\sun &\textbf{0.96} &0.87 &0.55 &\textbf{0.96} \\ \midrule[.5pt]\midrule
&&\\[-5pt]
\textbf{Average}& &\textbf{0.82}& 0.40& 0.66& \textbf{0.82}
\\ \bottomrule
\end{tabular}
}
\caption{Overview of results for transferability for target selection in image classification, where the transferability of subsampled target datasets are estimated (following the setup in~\cite{nguyen2020leep}). From the results we observe that the proposed GBC method performs on par with the current state-of-the-art LEEP method.}
\label{tab:target-selection}
\end{table}%

\begin{figure*}[t]
\hspace{-.6cm}
\includegraphics[width=1.07\textwidth]{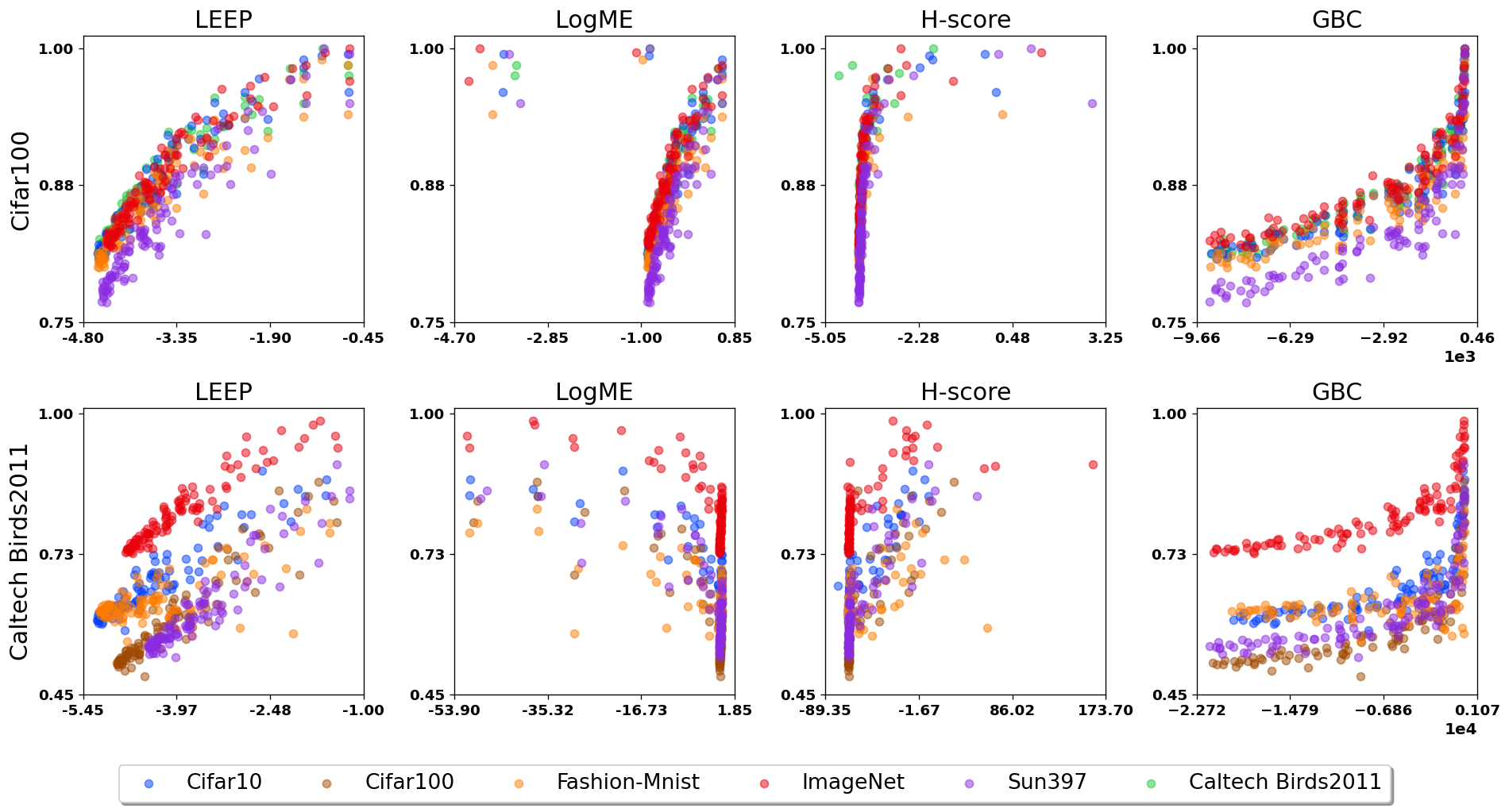}
\caption{This figure illustrates the scatter plots of LEEP, LogME, H-score, and GBC for CIFAR-100 and CUB'11 as target datasets. In each figure, the transferability score $\mathcal{S}_{s \rightarrow t}$ of the method is on the \textit{X-axis}, with the corresponding $\mathcal{A}_{s \rightarrow t}$ of each fine-tuned model on the \textit{Y-axis}. From the plots we observe that while LogME and H-score tend to struggle to differentiate between target datasets, both GBC and LEEP showcase increasing trends.}
\label{fig:target_selection_scatter}
\end{figure*}

\subsection{Classification: dataset transferability}
\label{sec:target-selection}

\xhdr{Experimental setup}
Good transferability metrics should correlate with a model's performance on target test data, as mentioned in \autoref{sec:formal-description}. To evaluate this, we follow the setup from \cite{nguyen2020leep}: Given a fixed source model, the goal is to rank target datasets according to the actual performance of the source model after fine-tuning it on the target training set.

For this set of experiments all our source models have a ResNet-50~\cite{he16cvpr} architecture. Our first source model is trained on ImageNet~\cite{russakovsky15ijcv}.
This ImageNet~\cite{russakovsky15ijcv} source model also acts as initialisation for the other 5 source models, trained on the following datasets: CIFAR-10 \& CIFAR-100~\cite{krizhevsky09}, Fashion-MNIST~\cite{xiao2017fashionmnist}, SUN397~\cite{xiao10cvpr} and Caltech-USCD Birds2011~\cite{CUB2011}.
This results in 6 source models. We use the same datasets as targets except for ImageNet, resulting in 5 target datasets. This results in $25$ source-target pairs used as experiments.

For each of these 25 experiments, we use a single source model and a single \textit{main} target dataset. 
Following~\cite{nguyen2020leep}, we construct a set of $100$ \textit{subsampled} target datasets from this main target dataset.
Each subsampled target dataset is obtained by sampling uniformly between $2\%$ and a $100\%$ of the target classes and using all available images for these classes.
For example, when the CIFAR-100 dataset is used as main target dataset, $100$ subsampled datasets are created, each containing the CIFAR-100 images for 2--100 (randomly selected) classes. 

For each of these subsampled target datasets, the transferability score of the source model is estimated.
To obtain the target model $m_{s\rightarrow t}$, we fully fine-tuned the source model for each subsampled target dataset, for 100 epochs (using SGD with Momentum, with a batch size of $10$ and learning rate of $10^{-3}$, these hyper-parameters are the same as in LEEP~\cite{nguyen2020leep}).

\xhdr{Evaluation}
The accuracy $\mathcal{A}_{s \rightarrow t}$ is obtained by evaluating the final target models on the target test set (removing labels not sampled for this particular target task). 
We measure correlation between the transferability metric $\mathcal{S}_{s \rightarrow t}$ and the accuracy $\mathcal{A}_{s \rightarrow t}$ using the weighted Kendall rank correlation $\tau_w$.  
The baselines we use are H-score, LogME, and LEEP.
For fair comparisons, each method is evaluated on the same set of $100$ random target datasets in all experiments.

\xhdr{Results}
We present the quantitative results in \autoref{tab:target-selection} (and more results in ~\autoref{app:image_classification}).
We observe that our proposed GBC has the top performance on $15$ (out of 25) experiments, LEEP has the top performance on $19$ experiments, LogME has the top performance in $5$ cases, and H-score has the top performance in $1$ case, where we include ties in our counting.

GBC and LEEP achieve an average $\tau_w$ score of $.82$, much higher than H-score ($.66$) and LogME ($.40$). 
Further, both GBC and LEEP consistently showcased high $\tau_w$ values ({$\geq .67$}) across all experiments, while both H-score ($.12$) and LogME underperform for certain target datasets (\eg CUB'11 for LogME).
These results confirm that the proposed GBC method outperforms LogME and H-score, and is on par with LEEP in this setting. 

We illustrate the correlation between $\mathcal{A}_{s\rightarrow t}$ and $\mathcal{S}_{s\rightarrow t}$ on CIFAR-100 (\textit{top}) and CUB'11 (\textit{bottom}) in \autoref{fig:target_selection_scatter}. We observe that LogME and H-score fail to distinguish well between certain target datasets, \ie assign near identical transferability scores despite the differences in target accuracies. On the other hand, both LEEP and the proposed GBC distinguish between the target datasets well, with persistent monotonically increasing trends.

\subsection{Segmentation: dataset transferability}
\newcommand{\semanticsegmentationdatasets}{
ADE20K~\cite{zhou17cvpr},
BDD~\cite{yu20cvpr},
CamVid~\cite{brostow09prl},
CityScapes~\cite{cordts16cvpr},
COCO~\cite{caesar18cvpr, lin14eccv, kirillov19cvpr}, 
IDD~\cite{varma19wacv},
iSAID~\cite{waqas2019isaid, xia18cvpr},
ISPRS~\cite{rottensteiner14isprs},
KITTI~\cite{alhaija18ijcv},
Mappilary~\cite{neuhold17cvpr},
Pascal Context~\cite{mottaghi14cvpr}, 
Pascal VOC~\cite{pascal-voc-2012}, 
ScanNet~\cite{dai17cvpr},
SUIM~\cite{islam2020suim},
SUN RGB-D~\cite{song15cvpr},
vGallery~\cite{weinzaepfel19cvpr}, and
vKITTI2~\cite{cabon20arxiv, gaidon16cvpr}}

\label{sec:semseg}

\xhdr{Experimental setup}
We now turn to a transfer learning scenario for semantic segmentation, following the setup of~\cite{mensink21arxiv}: 17 datasets spanning very different image domains (consumer photos, autonomous driving, aerial imagery, underwater, indoor scenes, synthetic, close-ups) containing 6--150 classes each:
\semanticsegmentationdatasets{}.
While~\cite{mensink21arxiv} used their setup to
investigate what factors are important for good transfer learning, they did not aim to predict transferability. Nevertheless, we interpret one of their measurements as a transferability metric.

We use the low-shot target training regime of~\cite{mensink21arxiv}, which is arguably the most interesting scenario for transfer learning. The target training set is limited to 150 images for all datasets, except COCO and ADE20k, where the limit is set to 1000 images since they contain a large number of classes.

We use a HRNetV2-W48 backbone~\cite{wang20pami} with a linear classifier on top. This model offers excellent performance for semantic segmentation~\cite{wang20pami} and was also used in~\cite{lambert20cvpr, mensink21arxiv}.
We train a source model on each dataset.

We consider all $17 \times 16 = 272$ valid $\langle$source model, target dataset$\rangle$ pairs (for each target dataset we do not consider its corresponding source model trained on the full training set).
For each pair, we compute the transferability metrics. We also compute the actual mean Intersection-over-Union performance
by fine-tuning the source model on the target training set, and then evaluate on the target test set.

We evaluate in two scenarios like before:
(1) given a fixed source model, we rank all valid target datasets;
and (2) given a fixed target dataset, we rank all valid source models.
For each scenario we measure the correlation with $\tau_w$ and also the top-1 selection accuracy:
for scenario (1) the percentage of targets where the source with the highest predicted transferability score also has the highest actual performance, and for scenario (2) the same, however with the role of source and target reversed.

\xhdr{GBC estimation}
For semantic segmentation, instead of one label per image, we have predictions at the pixel level. To estimate the transferability metrics, we consider each pixel $x_i$ and its ground truth label $y_i$ as a separate observation. 
Since using all observations for all metrics is too computationally expensive, we subsample 1000 pixels as observations per training image. We subsample using a class-balanced sampling strategy (\ie sample inverse-proportionally to the label frequency), which we found to improve results for all metrics. To make the comparison completely fair, we always use the exact same subsampled pixels for each image to calculate all transferability metrics.

For semantic segmentation, even after subsampling, we have generally many more observations than for image classification. Therefore, instead of modelling spherical Gaussians, we model Gaussians with a diagonal covariance matrix, which offer a greater modeling capacity.
This improved results for our method.

\xhdr{Image Domain Similarity} 
In~\cite{mensink21arxiv} they demonstrated that transfer learning performance was reasonably correlated with image domain similarity (IDS) between the source and target dataset. In this paper we interpret IDS as a transferability metric. IDS was established as follows~\cite{mensink21arxiv}: First a multi-task model (trained on multiple sources) was applied to 1000 randomly sampled images of each dataset, resulting in a single embedding vector per image. Then each target image embedding was matched to its closest source image embedding. Finally, IDS is the average euclidean distance between these matched embeddings. We obtained all IDS metrics from the authors in personal correspondence.

\xhdr{Results}
\autoref{tab:semantic_segmentation} presents the results (see \autoref{app:segmentation} for more details).
For scenario (1), when choosing a source model for a fixed target dataset,
our method has the highest top-1 accuracy: it outperforms all other methods in choosing the best source, which is the main goal in a practical application. When looking at the weighted Kendall $\tau_w$, which measures overall ranking correctness, LogME is best, and our method is second.
In scenario (2), determining for a fixed source model to which target dataset it transfers best, LogME completely fails. While IDS and LEEP perform better, they are still significantly below our method in both top-1 accuracy and $\tau_w$.
We conclude that our proposed GBC transferability metric is the overall best transferability metric for semantic segmentation.

\begin{table}[t]
    \centering
    \vspace{0.2cm}
    \resizebox{\columnwidth}{!}{
        \begin{tabular}{llcccc}
        \toprule
                                        &        & IDS   & LEEP                  & LogME                 & GBC     \\
                                        &    & \cite{mensink21arxiv}       & \cite{nguyen2020leep} & \cite{you2021logme}   & \emph{Ours} \\
        \midrule
        \multirow{2}{*}{fixed target}   & top-1   & 0.41   & 0.47                  & 0.59                  & \bf{0.65}   \\
        \cmidrule{2-6}
                                        & $\tau_w$ & 0.45 & 0.53                  & \bf{0.63}             & 0.59   \\
        \midrule                                                                        
        \multirow{2}{*}{fixed source}   & top-1     & 0.41 & 0.24                  & 0.00                  & \bf{0.76}   \\
        \cmidrule{2-6}
                                        & $\tau_w$ & 0.36 & 0.62                  & 0.08                  & \bf{0.69}   \\
        \bottomrule
        \end{tabular}
    }\vspace{-1mm}
    \caption{Overview of results for transferability estimation for semantic segmentation. %
    GBC outperforms all transferability methods in terms of top-1 accuracy, which is the most important measure in a practical transfer learning application.} 
    \vspace{-3mm}
    \label{tab:semantic_segmentation}
\end{table}

\section{Conclusion}

In this paper, we introduce the Gaussian Bhattacharyya coefficient (GBC), a novel transferability metric which measures the amount of overlap between target classes (each modelled as a Gaussian) in the source feature space.
The societal impact is that it reduces the need for heavy training procedures in transfer learning by selecting good models to transfer from in an efficient manner.
We compare our method against state-of-the-art transferability metrics: LogME~\cite{you2021logme}, LEEP~\cite{nguyen2020leep}, and H-score~\cite{bao19icip} and show that GBC outperforms them (or is on par) on most evaluation criteria. 
A key limitation of GBC is that it is designed for classification tasks only (not regression).

\appendix
\section{Limitations and Future work}
Here we reflect on some of the key limitations of the proposed GBC method.%

\paragraph{Network Architectures.}
Firstly, the selected source architectures play a key role in evaluating the proposed method. It is plausible that different architectures yield different class distributions in the embedding space,  which could impact the per-class Gaussian approximation of GBC. Further, all the architectures are sensitive to training hyperparameter choices, which introduces further complications in estimating ground truth transferability scores. To allow for fair comparison, we have used the same network architectures as in~\cite{you2021logme} as much as possible, and verified that our networks are trained until convergence. However, a different set of used architectures may influence the results.

\paragraph{Classification networks.}
GBC measures pairwise class overlap and hence is suitable for transfer in a classification setting. However, many interesting transfer learning problems are in regression, or even unsupervised learning and reinforcement learning. In terms of regression, it would be useful to extend GBC, for example by binning the regression variable and replace the \emph{class} overlap with an overlap between the used bins. We believe all of these directions present promising avenues for future work.

\section{Empirical analysis of design choices}
\label{app:ablations}

\begin{figure}[t]
    \centering
    \begin{minipage}[t]{.45\textwidth}
        \centering
        \includegraphics[width=.65\textwidth]{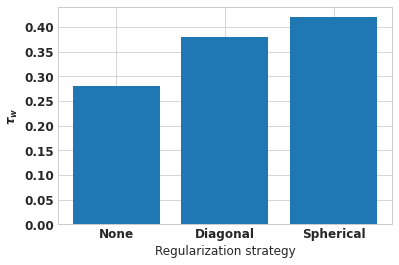}
        \caption{This figure demonstrates the superiority of spherical regularization with respect to other strategies in terms of weighted Kendall's rank correlation.}
        \label{fig:ablation_reg}
    \end{minipage}
    \qquad
    \begin{minipage}[t]{.45\textwidth}
        \centering
        \includegraphics[width=0.65\textwidth]{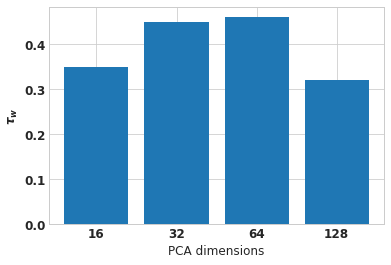}
        \caption{This figure demonstrates the sensitivity of GBC to PCA dimensionality on Cifar10 in terms of weighted Kendall's rank correlation.}
        \label{fig:ablation_pca}
    \end{minipage}
\end{figure}

In this experiment, we evaluate the influence of GBC's design choices introduced in our method section. We use CIFAR10 as a target dataset and transfer from the 9 source architectures pre-trained on ImageNet as described in our experiments.

\paragraph{Effects of regularization strategies} We compare three Gaussian covariance regularization strategies: none, diagonal, and spherical. As we can observe in \autoref{fig:ablation_reg}, adding regularization improves our results, with the best $\tau_w$ in the case of spherical regularization.

\paragraph{Effects of PCA dimensions} In this experiment, we compare the performance of GBC across multiple PCA dimensions: 16, 32, 64, and 128. We discover performance improvements up to 64 dimensions (see \autoref{fig:ablation_pca}), after which we observe a decline in $\tau_w$. Hence, it can be beneficial to carefully select the appropriate PCA dimensions for the given use case.

\mathchardef\mhyphen="2D
\newcommand{\Y}{$\checkmark$}
\newcommand{\N}{$\mhyphen$}

\begin{table*}[p]
  \centering
  \begin{minipage}[t]{.75\textwidth}
  \resizebox{\textwidth}{!}{%
  \begin{tabular}{c|*{4}{@{\hspace{3.5mm}}c}|*{4}{@{\hspace{3.5mm}}c}}
    \toprule
     & &\multicolumn{2}{c}{$\tau_w$} && &\multicolumn{2}{c}{Top-1}&\\
    \textbf{Target Dataset} & \textbf{IDS} & \textbf{LEEP} & \textbf{LogME} & \textbf{GBC} & \textbf{IDS} & \textbf{LEEP} & \textbf{LogME} & \textbf{GBC} \\
     & \cite{mensink21arxiv}& \cite{nguyen2020leep} & \cite{you2021logme}   & \textit{Ours}   & \cite{mensink21arxiv}&  \cite{nguyen2020leep} & \cite{you2021logme}   & \textit{Ours} \\
    \midrule
    Pascal Context~\cite{mottaghi14cvpr} & 0.48 & 0.64 &\textbf{ 0.82} & 0.73 & \N & \Y & \Y & \Y\\
    Pascal VOC~\cite{pascal-voc-2012} & 0.25 & 0.29 &\textbf{ 0.52} & 0.44 & \N & \N & \Y & \Y
\\
    ADE20K~\cite{zhou17cvpr} & 0.21 & 0.42 & \textbf{0.43} & 0.41 & \Y & \Y & \Y & \Y
\\
    COCO~\cite{caesar18cvpr, lin14eccv, kirillov19cvpr} & -0.12 & -0.15 & \textbf{0.20} & -0.14 & \N & \N & \N & \N
\\
    KITTI~\cite{alhaija18ijcv} & 0.64 & 0.77 & 0.82 & \textbf{0.84} & \N & \N & \N & \N
\\
    CamVid~\cite{brostow09prl} & 0.62 & \textbf{0.76} & 0.75 & 0.73 & \Y & \Y & \Y & \Y
\\
    CityScapes~\cite{cordts16cvpr} & 0.62 & 0.88 & 0.92 &\textbf{ 0.93} & \N & \Y & \Y & \Y
\\
    IDD~\cite{varma19wacv} & 0.57 & 0.80 & 0.87 & \textbf{0.90} & \Y & \Y & \Y & \Y
\\
    BDD~\cite{yu20cvpr} & 0.77 & 0.85 & \textbf{0.93 }& 0.90 & \Y & \Y & \Y & \Y
\\
    MVD~\cite{neuhold17cvpr} & 0.58 & 0.65 & \textbf{0.72 }& 0.66 & \N & \N & \N & \N
\\
    ISPRS~\cite{rottensteiner14isprs} & 0.26 & 0.08 & 0.52 & \textbf{0.66} & \Y & \N & \N & \Y
\\
    iSAID~\cite{waqas2019isaid, xia18cvpr} & 0.35 & \textbf{0.36} & 0.13 & 0.27 & \N & \N & \Y & \Y
\\
    SUN RGB-D~\cite{song15cvpr}  & 0.53 & 0.52 & \textbf{0.61} & 0.57 & \Y & \N & \N & \N
\\
    ScanNet~\cite{dai17cvpr} & 0.43 & \textbf{0.71} & 0.65 & 0.61 & \Y & \N & \N & \N
\\
    SUIM~\cite{islam2020suim} & 0.39 & 0.27 & \textbf{0.64} & 0.58 & \N & \Y & \Y & \Y
\\
    vKITTI2~\cite{cabon20arxiv, gaidon16cvpr} & 0.70 & 0.65 & \textbf{0.73 }& 0.64 & \N & \Y & \Y & \Y
\\
    vGallery~\cite{weinzaepfel19cvpr} & 0.40 & 0.44 &\textbf{ 0.50} & 0.27 & \N & \N & \N & \N
\\
    \bottomrule
    Average & 0.45 & 0.53 & \bf{0.63}  & 0.59 & 0.41   & 0.47    & 0.59   & \bf{0.65}  \\
  \end{tabular}
  }
  \caption{Per-target results for segmentation source selection.}
  \label{tab:source-selection-segmentation}
  \end{minipage}
  
  \begin{minipage}[t]{.75\textwidth}
  \resizebox{\textwidth}{!}{%
  \begin{tabular}{c|*{4}{@{\hspace{3.5mm}}c}|*{4}{@{\hspace{3.5mm}}c}}
    \toprule
     & &\multicolumn{2}{c}{$\tau_w$} && &\multicolumn{2}{c}{Top-1}&\\
    \textbf{Source Dataset} & \textbf{IDS} & \textbf{LEEP} & \textbf{LogME} & \textbf{GBC} & \textbf{IDS} & \textbf{LEEP} & \textbf{LogME} & \textbf{GBC} \\
      & \cite{mensink21arxiv}& \cite{nguyen2020leep} & \cite{you2021logme}   & \textit{Ours}   & \cite{mensink21arxiv}&  \cite{nguyen2020leep} & \cite{you2021logme}   & \textit{Ours} \\
    \midrule
    Pascal Context~\cite{mottaghi14cvpr} & 0.53 & \textbf{0.78} & 0.01 & 0.72 & \Y & \N & \N & \N
\\
    Pascal VOC~\cite{pascal-voc-2012} & 0.34 & 0.52 & 0.06 & \textbf{0.69} & \Y & \N & \N & \Y
\\
    ADE20K~\cite{zhou17cvpr} & 0.48 & \textbf{0.74} & 0.01 & 0.69 & \N & \Y & \N & \Y
\\
    COCO~\cite{caesar18cvpr, lin14eccv, kirillov19cvpr} & 0.12 & 0.66 & 0.15 & \textbf{0.73} & \N & \N & \N & \N
\\
    KITTI~\cite{alhaija18ijcv} & 0.48 & 0.60 & 0.09 & \textbf{0.70} & \N & \N & \N & \Y
 \\
    CamVid~\cite{brostow09prl} & 0.61 & 0.57 & 0.01 &\textbf{ 0.72} & \N & \N & \N & \Y
\\
    CityScapes~\cite{cordts16cvpr} & 0.38 & 0.57 & 0.17 & \textbf{0.74} & \N & \N & \N & \Y
\\
    IDD~\cite{varma19wacv} & 0.48 & 0.59 & 0.24 & \textbf{0.70} & \N & \N & \N & \Y
\\
    BDD~\cite{yu20cvpr} & 0.55 & 0.69 & 0.12 & \textbf{0.74} & \Y & \N & \N & \Y
\\
    MVD~\cite{neuhold17cvpr} & 0.58 & 0.63 & 0.21 & \textbf{0.68} & \N & \N & \N & \N
\\
    ISPRS~\cite{rottensteiner14isprs}  & -0.10 & 0.59 & 0.09 &\textbf{ 0.69 }& \N & \N & \N & \Y
 \\
    iSAID~\cite{waqas2019isaid, xia18cvpr} & 0.30 & 0.73 & 0.02 & \textbf{0.79} & \Y & \Y & \N & \Y
\\
    SUN RGB-D~\cite{song15cvpr} & 0.40 & 0.54 & 0.06 & \textbf{0.62 }& \Y & \Y & \N & \Y
\\
    ScanNet~\cite{dai17cvpr} & 0.40 & 0.59 & 0.05 &\textbf{ 0.60} & \Y & \Y & \N & \Y
\\
    SUIM~\cite{islam2020suim} & 0.32 & 0.61 & 0.11 &\textbf{ 0.72 }& \N & \N & \N & \Y
\\
    vKITTI2~\cite{cabon20arxiv, gaidon16cvpr} & 0.61 & 0.62 & 0.17 & \textbf{0.71} & \Y & \N & \N & \Y
\\
    vGallery~\cite{weinzaepfel19cvpr} & -0.01 & \textbf{0.52 }& -0.19 & 0.51 & \N & \N & \N & \N
\\
    \bottomrule
    Average & 0.36 & 0.62  & 0.08    & \bf{0.69} & 0.41 & 0.24    & 0.00      & \bf{0.76} \\
  \end{tabular}
  }
  \caption{Per-source results for segmentation target selection.}
  \label{tab:target-selection-segmentation}
  \end{minipage}
\end{table*}

\section{Detailed Results for Semantic Segmentation}
\label{app:segmentation}
Here we provide additional results for semantic segmentation. 
In particular \autoref{tab:source-selection-segmentation} shows per-target results for the source selection task (ranking source datasets for a particular fixed target dataset), while \autoref{tab:target-selection-segmentation} shows per-source results for the target selection task (ranking target datasets for a fixed source dataset). For each table we provide the Weighted Kendall Tau correlation metric for every dataset, and we indicate whether a transferability metric correctly predicts the top-1 source (in source selection) or the top-1 target (in target selection).
\section{Additional Results for Image Classification}
\label{app:image_classification}
\begin{figure*}[t]
 \centering
 \includegraphics[width=.95\textwidth]{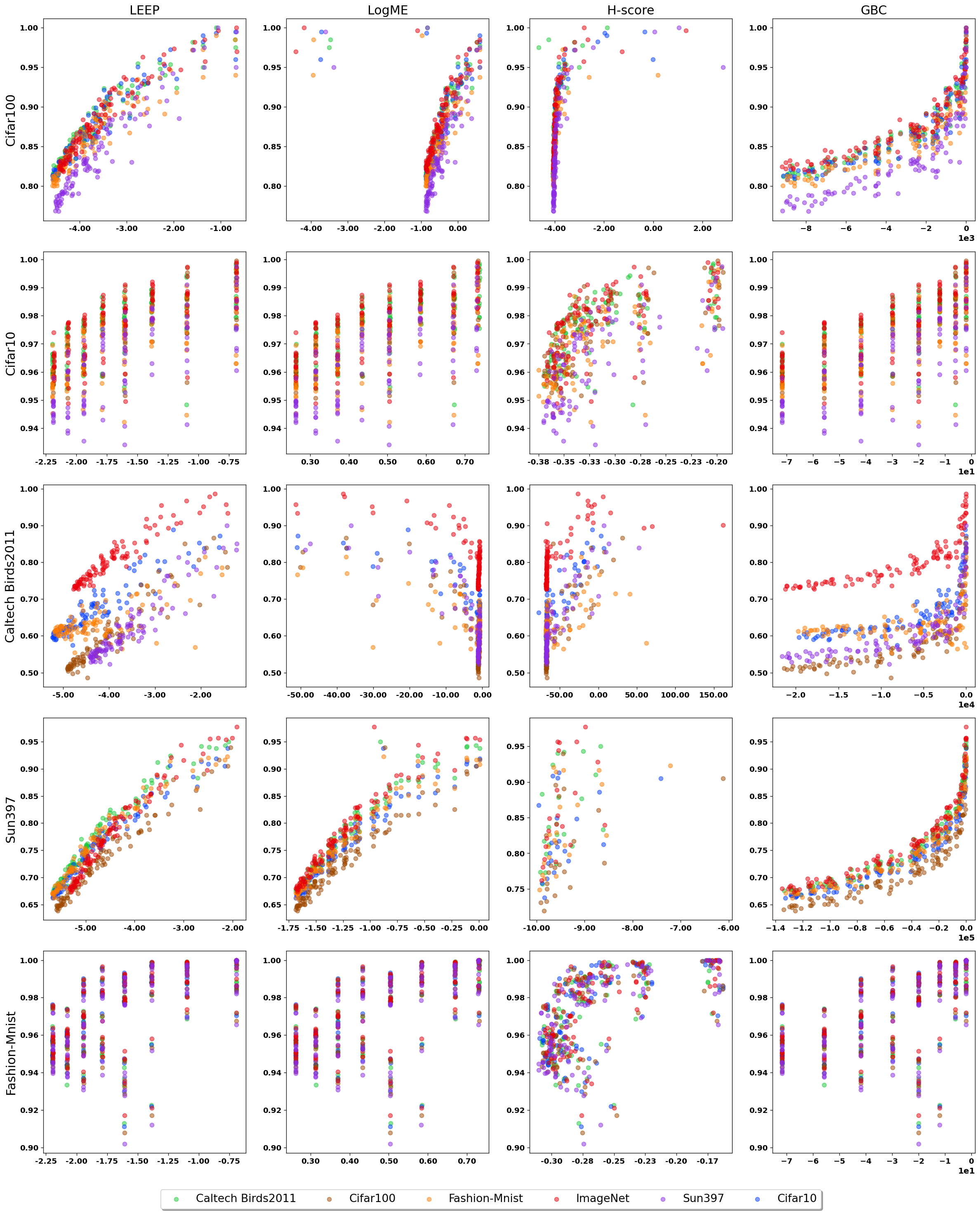}
 \caption{
 This figure illustrates the scatter plots of LEEP, LogME, H-score, and GBC for all datasets used in the dataset transferability experiment (see Section 4.2 and Fig 4 of the main paper).
 In each plane, the transferability score $\mathcal{S}_{s \rightarrow t}$ of the method is on the X-axis, with the corresponding $\mathcal{A}_{s \rightarrow t}$ of each fine-tuned model on the Y-axis. From the plots we observe that while LogME and H-score tend to struggle to differentiate between some of the target datasets, both GBC and LEEP showcase increasing trends.
 }
 \label{fig:appendix_target_selection_scatter_plots}
\end{figure*}

In this section we provide additional experimental results for the image classification experiments.
\begin{itemize}
    \item For the source selection experiments (\autoref{sec:source-selection} / \autoref{tab:source_selection_full}), we include different correlation metrics: Weighted Kendall Tau (\autoref{tab:appendix_source_selection_tau_weighted}, also in the main paper), Kendall Tau (\autoref{tab:appendix_source_selection_tau}), and Pearson's $r$ coefficient (\autoref{tab:appendix_source_selection_pearson}).
    \item For the dataset transferability experiments (\autoref{sec:target-selection} / \autoref{tab:target-selection}),
    we include also the
    Weighted Kendall Tau (\autoref{tab:appendix_target_selection_tau_weighted}, used in the main paper), 
    Kendall Tau (\autoref{tab:appendix_target_selection_tau}), and 
    Pearson's $r$ coefficient (\autoref{tab:appendix_target_selection_pearson}).
    \item Moreover, we provide scatter plots for all target datasets used in~\autoref{fig:appendix_target_selection_scatter_plots}, extending Figure 4 in the main paper.
\end{itemize}

\begin{table*}[p]
    \begin{subtable}[t]{.99\textwidth}
        \centering
        \resizebox{.85\textwidth}{!}{{%
\begin{tabular}{ccccccccc|c} \toprule
& Pets & Imagenette& CIFAR-10& CUB'11& Dogs & Flowers102& SUN & CIFAR-100 &\textbf{Average}\\ \midrule
LogMe &-0.06& 0.58& 0.25& 0.2& 0.08& 0.00& -0.19& 0.34 &0.15\\
H-score &0.06& 0.59& 0.45& 0.16& -0.01& 0.09& 0.09& 0.34 &0.22\\
LEEP &\textbf{0.63}& \textbf{0.65}& \textbf{0.52}& 0.25& 0.59& -0.46& \textbf{0.40} & \textbf{0.55} &0.39 \\
GBC &0.55& 0.63& 0.46& \textbf{0.43}& \textbf{0.80}& \textbf{0.23}& 0.32& 0.35 &\textbf{0.47}\\ \bottomrule
\end{tabular}
}}
        \caption{Metric: Weighted Kendall Tau ($\tau_w$)}
        \label{tab:appendix_source_selection_tau_weighted}
    \end{subtable}\vspace{1mm}
    
    \begin{subtable}[t]{.99\textwidth}
        \centering
        \resizebox{.85\textwidth}{!}{{%
\begin{tabular}{ccccccccc|c} \toprule
& Pets & Imagenette& CIFAR-10& CUB'11& Dogs & Flowers102& SUN & CIFAR-100 &\textbf{Average}\\ \midrule
LogME &-0.06& \textbf{0.56}& 0.44& \textbf{0.28}& 0.17& 0.0& -0.17& 0.50 &0.22\\
H-score &0.17& 0.54& \textbf{0.56}& \textbf{0.28}& 0.06& 0.13& 0.0& 0.50 &0.28\\
LEEP &\textbf{0.5}& 0.5& \textbf{0.56}& \textbf{0.28}& 0.39& -0.28& \textbf{0.28}& \textbf{0.56} &0.35 \\
GBC &0.44& 0.39& 0.50& 0.22& \textbf{0.67}& \textbf{0.17}& 0.17& 0.39 &\textbf{0.37}\\\bottomrule
\end{tabular}
}}
        \caption{Metric: Kendall Tau ($\tau$)}
        \label{tab:appendix_source_selection_tau}
    \end{subtable}\vspace{1mm}
    
    \begin{subtable}[t]{.99\textwidth}
        \centering
        \resizebox{.85\textwidth}{!}{{%
\begin{tabular}{ccccccccc|c} \toprule
& Pets & Imagenette& CIFAR-10& CUB'11& Dogs & Flowers102& SUN & CIFAR-100 &\textbf{Average}\\ \midrule
LogME &0.33& \textbf{0.58}& 0.62& 0.45& 0.28& 0.37& -0.14& 0.54 &0.38\\
H-score &0.37& 0.52& \textbf{0.83}& 0.35& 0.25& -0.0& -0.01& 0.78 &0.39\\
LEEP &0.28& 0.12& 0.81& 0.03& 0.48& -0.2& 0.21& \textbf{0.75} &0.31 \\
GBC &\textbf{0.40} & 0.45& 0.81& \textbf{0.49}& \textbf{0.44}& \textbf{0.57}& \textbf{0.27}& \textbf{0.77} &\textbf{0.52}\\
 \bottomrule
\end{tabular}
}}
        \caption{Metric: Pearson correlation coefficient ($\rho$)}
        \label{tab:appendix_source_selection_pearson}
    \end{subtable}\vspace{-1mm}
    \caption{
    Overview of results for transferability for source selection in image classification.
    }
    \label{tab:appendix_source_selection}
\end{table*}

\begin{table*}[p]
    \centering
    \begin{subtable}[t]{.35\textwidth}
        \centering
        \resizebox{!}{60mm}{{%
\begin{tabular}{cHcccc} \toprule
 \textbf{Source} & \textbf{Target dataset}  & \textbf{LEEP}         & \textbf{LogME}        & \textbf{H-score}      & \textbf{GBC}\\ 
                 &                          & \cite{nguyen2020leep} & \cite{you2021logme}   & \cite{bao2019hscore}  & \textit{Ours}\\\midrule
&&\multicolumn{4}{c}{\textbf{CIFAR-10}}\\
\birds &\cifar &0.68 &0.71 &0.69 &\textbf{0.72} \\
\cifarh &\cifar &\textbf{0.75} &\textbf{0.75} &0.73 &0.69 \\
\fashion &\cifar &0.68 &0.70 &\textbf{0.72} &0.68 \\
\sun &\cifar &0.73 &\textbf{0.75} &0.72 &0.67 \\ 
ImageNet &\cifar &0.68 &0.68 &0.69 &\textbf{0.71} \\\midrule
&&\multicolumn{4}{c}{\textbf{CIFAR-100}}\\
\birds &\cifarh &\textbf{0.90} &0.29 &0.59 &\textbf{0.90} \\
\cifar &\cifarh &\textbf{0.92} &0.29 &0.88 &\textbf{0.92} \\
\fashion &\cifarh &\textbf{0.88} &0.24 &0.26 &\textbf{0.88} \\
\sun &\cifarh &\textbf{0.90} &0.30 &0.88 &\textbf{0.90} \\ 
ImageNet &\cifarh &0.91 &0.25 &0.88 &\textbf{0.92} \\\midrule
&&\multicolumn{4}{c}{\textbf{Fashion-MNIST}}\\
\birds &\fashion &\textbf{0.72} &0.71 &0.71 &0.71 \\
\cifar &\fashion &0.72 &\textbf{0.73} &0.69 &0.69 \\
\cifarh &\fashion &\textbf{0.71} &\textbf{0.71} &0.70 &0.69 \\
\sun &\fashion &\textbf{0.71} &\textbf{0.71} &0.69 &\textbf{0.71} \\ 
ImageNet &\fashion &\textbf{0.72} &0.71 &0.69 &0.70 \\\midrule
&&\multicolumn{4}{c}{\textbf{Caltech-USCD Birds 2011}}\\
\cifar &\birds &\textbf{0.87} &-0.59 &0.83 &0.86 \\
\cifarh &\birds &\textbf{0.87} &-0.58 &0.80 &\textbf{0.87} \\
\fashion &\birds &\textbf{0.70} &-0.50 &0.51 &0.69 \\
\sun &\birds &\textbf{0.88} &-0.60 &0.80 &\textbf{0.88} \\
ImageNet &\birds &\textbf{0.89} &-0.59 &0.73 &0.88 \\\midrule
&&\multicolumn{4}{c}{\textbf{SUN-397}}\\
\birds &\sun &\textbf{0.95} &0.87 &0.54 &\textbf{0.95} \\
\cifar &\sun &\textbf{0.95} &0.87 &0.12 &\textbf{0.95} \\
\cifarh &\sun &\textbf{0.95} &0.88 &0.51 &\textbf{0.95} \\
\fashion &\sun &\textbf{0.95} &0.86 &0.54 &\textbf{0.95} \\
ImageNet &\sun &\textbf{0.96} &0.87 &0.55 &\textbf{0.96} \\ \midrule[.5pt]\midrule
&&\\[-5pt]
\textbf{Average}& &\textbf{0.82}& 0.40& 0.66& \textbf{0.82}
\\ \bottomrule
\end{tabular}
}}
        \caption{Weighted Kendall Tau ($\tau_w$)}
        \label{tab:appendix_target_selection_tau_weighted}
    \end{subtable}
    \begin{subtable}[t]{.27\textwidth}
        \centering
        \resizebox{!}{60mm}{{%
\begin{tabular}{HHcccc} \toprule
 \textbf{Source} & \textbf{Target dataset}  & \textbf{LEEP}         & \textbf{LogME}        & \textbf{H-score}      & \textbf{GBC}\\ 
                 &                          & \cite{nguyen2020leep} & \cite{you2021logme}   & \cite{bao2019hscore}  & \textit{Ours}\\\midrule
&&\multicolumn{4}{c}{\textbf{CIFAR-10}}\\
\birds &\cifar  &0.51 &0.53 &0.54 &\textbf{0.55} \\
\cifarh &\cifar &\textbf{0.58} &0.57 &0.56 &0.55 \\
\fashion &\cifar &0.49 &0.48 &\textbf{0.50} &0.49 \\
ImageNet &\cifar &0.53 &0.55 &\textbf{0.56} &\textbf{0.56} \\
\sun &\cifar &\textbf{0.53} &\textbf{0.53} &0.50 &0.49 \\ \midrule
&&\multicolumn{4}{c}{\textbf{CIFAR-100}}\\
\birds &\cifarh &\textbf{0.84} &0.74 &0.65 &\textbf{0.84} \\
\cifar &\cifarh &\textbf{0.85} &0.74 &0.73 &\textbf{0.85} \\
\fashion &\cifarh &\textbf{0.81} &0.69 &0.66 &\textbf{0.81} \\
\sun &\cifarh &\textbf{0.83} &0.72 &0.74 &0.82 \\ 
ImageNet &\cifarh &\textbf{0.86} &0.74 &0.77 &\textbf{0.86} \\\midrule
&&\multicolumn{4}{c}{\textbf{Fashion-MNIST}}\\
\birds &\fashion &0.48 &0.46 &\textbf{0.49} &0.48 \\
\cifar &\fashion &0.47 &\textbf{0.49} &0.46 &0.46 \\
\cifarh &\fashion &0.47 &\textbf{0.48} &0.46 &0.46 \\
\sun &\fashion  &\textbf{0.49} &0.47 &0.47 &0.47 \\ 
ImageNet &\fashion&0.49 &0.48 &0.47 &\textbf{0.50} \\\midrule
&&\multicolumn{4}{c}{\textbf{Caltech-USCD Birds 2011}}\\
\cifar &\birds &\textbf{0.79} &-0.04 &0.69 &\textbf{0.79} \\
\cifarh &\birds  &0.79 &-0.02 &0.67 &\textbf{0.80} \\
\fashion &\birds  &\textbf{0.33} &-0.03 &0.29 &\textbf{0.33} \\
\sun &\birds &\textbf{0.79} &-0.04 &0.67 &0.78 \\
ImageNet &\birds  &\textbf{0.82} &-0.02 &0.65 &\textbf{0.82} \\\midrule
&&\multicolumn{4}{c}{\textbf{SUN-397}}\\
\birds &\sun &\textbf{0.92} &0.91 &0.27 &\textbf{0.92} \\
\cifar &\sun &\textbf{0.92} &0.90 &0.22 &0.91 \\
\cifarh &\sun &\textbf{0.90} &0.89 &0.28 &\textbf{0.90} \\
\fashion &\sun &\textbf{0.91} &0.90 &0.26 &\textbf{0.91} \\
ImageNet &\sun &\textbf{0.92} &0.91 &0.26 &\textbf{0.92} \\ \midrule[.5pt]\midrule
&&\\[-5pt]
\textbf{Average}& &\textbf{0.69}& 0.52& 0.51& \textbf{0.69}
\\ \bottomrule
\end{tabular}
}}
        \caption{Kendall Tau ($\tau$)}
        \label{tab:appendix_target_selection_tau}
    \end{subtable}
    \begin{subtable}[t]{.27\textwidth}
        \centering
        \resizebox{!}{60mm}{{%
\begin{tabular}{HHcccc} \toprule
 \textbf{Source} & \textbf{Target dataset}  & \textbf{LEEP}         & \textbf{LogME}        & \textbf{H-score}      & \textbf{GBC}\\ 
                 &                          & \cite{nguyen2020leep} & \cite{you2021logme}   & \cite{bao2019hscore}  & \textit{Ours}\\\midrule
&&\multicolumn{4}{c}{\textbf{CIFAR-10}}\\
\birds &\cifar &0.69 &\textbf{0.70} &0.65 &\textbf{0.70} \\
\cifarh &\cifar &0.75 &\textbf{0.77} &0.70 &\textbf{0.77} \\
\fashion &\cifar &0.63 &0.65 &0.58 &\textbf{0.66} \\
ImageNet &\cifar &0.71 &\textbf{0.73} &0.66 &0.72 \\
\sun &\cifar &0.69 &\textbf{0.71} &0.64 &0.70 \\ \midrule
&&\multicolumn{4}{c}{\textbf{CIFAR-100}}\\
\birds &\cifarh &\textbf{0.94} &0.29 &0.56 &0.87 \\
\cifar &\cifarh &\textbf{0.95} &0.29 &0.61 &0.86 \\
\fashion &\cifarh &\textbf{0.92} &0.21 &-0.29 &0.85 \\
ImageNet &\cifarh &\textbf{0.95} &0.19 &0.56 &0.87 \\
\sun &\cifarh &\textbf{0.95} &0.25 &0.58 &0.82 \\\midrule
&&\multicolumn{4}{c}{\textbf{Fashion-MNIST}}\\
\birds &\fashion &0.63 &\textbf{0.64} &0.60 &0.61 \\
\cifar &\fashion &0.61 &\textbf{0.62} &0.57 &0.59 \\
\cifarh &\fashion &0.61 &\textbf{0.62} &0.56 &0.59 \\
ImageNet &\fashion &\textbf{0.63} &\textbf{0.63} &0.59 &0.60 \\
\sun &\fashion &0.62 &\textbf{0.63} &0.59 &0.60 \\\midrule
&&\multicolumn{4}{c}{\textbf{Caltech-USCD Birds 2011}}\\
\cifar &\birds &\textbf{0.94} &-0.75 &0.87 &0.77 \\
\cifarh &\birds &\textbf{0.94} &-0.77 &0.81 &0.76 \\
\fashion &\birds &\textbf{0.66} &-0.68 &0.24 &0.43 \\
ImageNet &\birds &\textbf{0.95} &-0.72 &0.56 &0.83 \\
\sun &\birds &\textbf{0.94} &-0.77 &0.76 &0.75 \\\midrule
&&\multicolumn{4}{c}{\textbf{SUN-397}}\\
\birds &\sun &\textbf{0.92} &0.91 &0.27 &\textbf{0.92} \\
\cifar &\sun &\textbf{0.92} &0.90 &0.22 &0.91 \\
\cifarh &\sun &\textbf{0.90} &0.89 &0.28 &\textbf{0.90} \\
\fashion &\sun &\textbf{0.91} &0.90 &0.26 &\textbf{0.91} \\
ImageNet &\sun &\textbf{0.92} &0.91 &0.26 &\textbf{0.92} \\ \midrule[.5pt]\midrule
&&\\[-5pt]
\textbf{Average}& &\textbf{0.82}& 0.36& 0.54& 0.75
\\ \bottomrule
\end{tabular}
}}
        \caption{Pearson correlation coefficient ($\rho$)}
        \label{tab:appendix_target_selection_pearson}
    \end{subtable}\vspace{-1mm}
    \caption{Overview of results for transferability for target dataset transferability.}
    \label{tab:appendix_target_selection}
\end{table*}

\clearpage
{\small
\bibliographystyle{ieee_fullname}
\bibliography{shortstrings,loco,loco_extra,loco_datasets}
}

\end{document}